\documentclass[letterpaper, 10 pt, conference]{ieeeconf}  

\IEEEoverridecommandlockouts                              

\usepackage{graphics} 
\usepackage[pdftex]{graphicx}
\usepackage{circuitikz}
\usepackage{tikz}
\usetikzlibrary{calc,patterns,decorations.pathmorphing,decorations.markings,decorations.pathreplacing}
\usepackage{tikzscale}
\usetikzlibrary{patterns,decorations.pathmorphing,positioning}
\usetikzlibrary{arrows}
\usepackage{scalefnt}
\graphicspath{ {./images/} }
\usepackage{amsmath}
\usepackage{float}
\usepackage{circuitikz}
\usepackage{pgfplots}
\usepgfplotslibrary{units}
\usepackage{filecontents}
\usepackage{tikz,pgfplots,filecontents,amsmath}
\usepackage{algorithmic}
\usepackage{amssymb}    

\usepackage{bm,times}
\usepackage{bm} 
\usepackage{color} 
\usepackage{hyperref}
\usepackage{cancel}
\usepackage[export]{adjustbox}
\usepackage{mathtools}
\usepackage{subfig}
\usepackage{caption}
\usepackage{subcaption}
\usepackage{comment}
\usepackage{fixltx2e}
\usepackage[capitalise]{cleveref}
\usetikzlibrary{calc}

\usepackage{todonotes}
\usepackage{pdfpages}
\usepackage{graphicx}
\usepackage{caption}
\usepackage{booktabs} 
\usepackage{amsmath,bm,mathtools,upgreek}
\usepackage{amssymb,stmaryrd}
\usepackage{pgfplots}
\usepackage{prettyref}

\newrefformat{fig}{Figure~[\ref{#1}]}
\newsavebox\IBoxA \newsavebox\IBoxB \newlength\IHeight
\newcommand\TwoFig[6]{
  \sbox\IBoxA{\includegraphics[width=0.3\textwidth]{#1}}
  \sbox\IBoxB{\includegraphics[width=0.3\textwidth]{#4}}%
  \ifdim\ht\IBoxA>\ht\IBoxB
    \setlength\IHeight{\ht\IBoxB}%
  \else\setlength\IHeight{\ht\IBoxA}\fi
  \begin{figure}[!htb]
  \minipage[t]{0.3\textwidth}\centering
  \includegraphics[height=\IHeight]{#1}
  \caption{#2}\label{#3}
  \endminipage\hfill
  \minipage[t]{0.3\textwidth}\centering
  \includegraphics[height=\IHeight]{#4}
  \caption{#5}\label{#6}
  \endminipage 
  \end{figure}%
}
\makeatletter 
\newcommand*{\b@xplus}[1][+]{\ooalign{%
    $\m@th\vcenter{\hbox{$\m@th#1$}}$\cr%
    \hidewidth$\m@th\boxempty$\hidewidth\cr}} 
\renewcommand*{\boxplus}{\mathbin{\b@xplus}} 
\renewcommand*{\boxminus}{\mathbin{\b@xplus[-]}} 
\makeatother

\title{\LARGE \bf
PACC: A Passive-Arm Approach for High-Payload Collaborative Carrying with Quadruped Robots Using Model Predictive Control}

\author{Giulio Turrisi$^{*1}$, Lucas Schulze$^{1,2}$, Vivian S. Medeiros$^{1,3}$, Claudio Semini$^{1}$ and Victor Barasuol$^{*1}$
\thanks{$^{1}$Authors are with Dynamic Legged Systems Lab, Istituto Italiano di Tecnologia (IIT). \tt\small name.lastname@iit.it.} 
\thanks{$^{2}$Author is with Intelligent Autonomous Systems Lab, TU Darmstadt.}
\thanks{$^{3}$Author is with Legged Robotics Group, Sao Paulo University (USP).}
\thanks{$^{*}$ Equal contribution.}
}

\usepackage{xcolor}
\usepackage{eso-pic}
\AddToShipoutPictureBG*{%
  \AtPageUpperLeft{%
    \setlength{\unitlength}{1mm}%
    \put(-9.5,-9){\makebox(\paperwidth,0)[c]{\parbox{0.9\textwidth}{2024 IEEE/RSJ International Conference on Intelligent Robots and Systems (IROS), pp. 11139-11146}}}
  }
}

\AddToShipoutPictureBG*{%
  \AtPageUpperLeft{%
    \setlength{\unitlength}{1mm}%
    \put(-9.5,-14){\makebox(\paperwidth,0)[c]{\parbox{0.9\textwidth}{DOI: \href{https://ieeexplore.ieee.org/document/10801456}{10.1109/IROS58592.2024.10801456}}}}
  }
}

\newcommand{\Real}{\mathbb{R}}
\newcommand{\Transp}{^\mathrm{\scriptscriptstyle T}}

\newcommand{\ZeroMat}{\mathbf{0}}

\newcommand{\CoMposlin}{\mathbf{r}}
\newcommand{\CoMposlinx}{\mathrm{r}_x}
\newcommand{\CoMposliny}{\mathrm{r}_y}
\newcommand{\CoMposlinz}{\mathrm{r}_z}
\newcommand{\CoMposlinScalar}{\mathrm{r}}
\newcommand{\CoMvellinx}{\dot{\mathrm{r}}_x}
\newcommand{\CoMvelliny}{\dot{\mathrm{r}}_y}

\newcommand{\CoMposang}{\boldsymbol{\Theta}}
\newcommand{\CoMvelang}{\boldsymbol{\omega}^B}
\newcommand{\dotCoMvelang}{\dot{\boldsymbol{\omega}}^B}
\newcommand{\gravity}{\mathrm{g}}
\newcommand{\gravityArray}{\mathbf{g}}
\newcommand{\Grf}{\mathbf{f}}

\newcommand{\contactPoint}{\mathbf{p}}
\newcommand{\massRobot}{\mathrm{m}}
\newcommand{\inertiaRobot}{\mathbf{I}}

\newcommand{\rotMatBTOW}{\mathbf{R}_B^{W}}
\newcommand{\rotMatWTOB}{\mathbf{R}_W^{B}}

\newcommand{\rotMatHTOW}{\mathbf{R}_H^{W}}
\newcommand{\TransOmegaToEuler}{W_{\eta}}
\newcommand{\ForceExt}{\mathbf{f}_\mathrm{ext}}
\newcommand{\TorqueExt}{\boldsymbol{\tau}_\mathrm{ext}}

\newcommand{\contactFlag}{\delta}
\newcommand{\contactFlagVector}{\boldsymbol{\delta}}

\newcommand{\FootPos}{\mathbf{p}}
\newcommand{\FootVel}{\dot{\mathbf{p}}}
\newcommand{\FootVelOpt}{\mathbf{v}}

\newcommand{\PAjoint}{\mathbf{q}_{a}}

\newcommand{\PAjointAcc}{\ddot{\mathbf{q}}_{a}}
\newcommand{\PAinertiaM}{\mathbf{M}_\mathrm{a}}
\newcommand{\PAV}{\mathbf{V}_\mathrm{a}}
\newcommand{\PAJee}{\mathbf{J}_\mathrm{ee}}
\newcommand{\PAFee}{\mathbf{f}_\mathrm{ee}}

\newcommand{\PAFeeH}{\mathbf{f}^H_\mathrm{ee}}
\newcommand{\PAFeeHx}{{\mathrm{f}^H_\mathrm{ee}}_x}
\newcommand{\PAFeeHy}{{\mathrm{f}^H_\mathrm{ee}}_y}
\newcommand{\PAFeeHz}{{\mathrm{f}^H_\mathrm{ee}}_z}

\newcommand{\PAPee}{\mathbf{p}_\mathrm{ee}}
\newcommand{\PAPeeHat}{\hat{\mathbf{p}}_\mathrm{ee}}
\newcommand{\PAPeex}{{\mathrm{p}_\mathrm{ee}}_x}
\newcommand{\PAPeey}{{\mathrm{p}_\mathrm{ee}}_y}
\newcommand{\PAPeez}{{\mathrm{p}_\mathrm{ee}}_z}
\newcommand{\PAPeeScalar}{\mathrm{p}_\mathrm{ee}}
\newcommand{\PAPeeH}{\mathbf{p}^H_\mathrm{ee}}
\newcommand{\PAPeeHx}{{\mathrm{p}^H_\mathrm{ee}}_x}
\newcommand{\PAPeeHy}{{\mathrm{p}^H_\mathrm{ee}}_y}
\newcommand{\PAPeeHz}{{\mathrm{p}^H_\mathrm{ee}}_z}

\newcommand{\PAtauG}{\boldsymbol{\tau}_\mathrm{g}}
\newcommand{\PAtauSpring}{\boldsymbol{\tau}_\mathrm{s}}
\newcommand{\PAtauDamper}{\boldsymbol{\tau}_\mathrm{d}}
\newcommand{\PASpringCoef}{\mathbf{k}_\mathrm{s}}
\newcommand{\PADamperCoef}{\mathbf{k}_\mathrm{d}}
\newcommand{\BarPASpringCoef}{\bar{\mathbf{k}}_\mathrm{s}}
\newcommand{\BarPASpringCoefScalar}{\bar{\mathrm{k}}_\mathrm{s}}

\newcommand{\KFFeeEst}{\hat{\mathbf{f}}_{ee}}

\newcommand{\xss}{{\mathbf{x}}}
\newcommand{\uss}{{\mathbf{u}}}

\newcommand{\dss}{{\mathbf{d}}}

\newcommand{\Ny}{\mathrm{n_y}}

\newcommand{\xref}{\xss^\mathrm{d}}
\newcommand{\Xopt}{\bar{\xss}}
\newcommand{\Uopt}{\bar{\uss}}

\newcommand{\Qfc}{\mathbf{Q}_\mathrm{x}}
\newcommand{\Rfc}{\mathbf{R}_\mathrm{u}}

\newcommand{\hFriction}{\mathbf{h}_\mathrm{fc}}
\newcommand{\ZMP}{\mathrm{zmp}}

\newcommand{\hZmp}{\mathbf{h}_\ZMP}

\newcommand{\xZMP}{x_\ZMP}
\newcommand{\yZMP}{y_\ZMP}

\newcommand{\FootPosNomXY}{\mathbf{p}^H_{xy}}
\newcommand{\FootPosNomZ}{\mathrm{p}^H_{z}}
\newcommand{\FootPosNomZI}{\mathrm{p}^H_{z_i}}
\newcommand{\FootPosNom}{\mathbf{p}_\mathrm{n}}

\newcommand{\FootPosNomHI}{\mathbf{p}^H_{\mathrm{n}_i}}

\newcommand{\FootPosCenter}{\mathbf{p}_\mathrm{c}}
\newcommand{\FootPosCenterH}{\mathbf{p}^H_{\mathrm{c}_i}}
\newcommand{\FootPosLiftOffZ}{{\mathrm{p}_\mathrm{lo}}_z}
\newcommand{\deltaFootPA}{\Delta_{pa}}

\newcommand{\YawVel}{\dot{\psi}}

\newcommand{\CoMAngRateDesired}{\dot{\CoMposang}_\mathrm{d}}

\newcommand{\UserVf}{\mathbf{V}_\mathrm{f}}
\newcommand{\UserVz}{\mathbf{V}_\mathrm{z}}

\newcommand{\UserVfxyH}{{\mathbf{V}_\mathrm{f}}_{xy}^H}

\newcommand{\DeltaSwingtime}{\Delta t_\mathrm{sw}}

\newcommand{\stepfreq}{f_\mathrm{{step}}}

\newcommand{\dutygait}{\mathrm{D}_\mathrm{f}}

\begin{document}

\maketitle
\thispagestyle{empty}
\pagestyle{empty}

\begin{abstract}
In this paper, we introduce the concept of using passive arm structures with intrinsic impedance for robot-robot and human-robot collaborative carrying with quadruped robots. The concept is meant for a leader-follower task and takes a minimalist approach that focuses on exploiting the robots' payload capabilities and reducing energy consumption, without compromising the robot locomotion capabilities. We introduce a preliminary arm mechanical design and describe how to use its joint displacements to guide the robot's motion. To control the robot's locomotion, we propose a decentralized Model Predictive Controller that incorporates an approximation of the arm dynamics and the estimation of the external forces from the collaborative carrying. We validate the overall system experimentally by performing both robot-robot and human-robot collaborative carrying on a stair-like obstacle and on rough terrain.


\end{abstract}

\section{INTRODUCTION}

Legged robots are becoming increasingly common as machines in our society.
They are versatile systems that can cover a wide range of tasks,
from domestic and urban to industrial applications, being able to navigate seamlessly
on flat and irregular surfaces, or in natural scenarios. So far, the
main focus of their application has been inspection and monitoring,
with minimal to no physical interaction with the environment or people.
In recent years, the research community has increased the effort to endow
legged platforms with manipulator arms or mechanical structures to make them perform physical interaction and object manipulation
\cite{kashiri19ral}, \cite{bellicoso19icra}, \cite{icra16brehman},
\cite{parosi23iros}, \cite{gamba23irim}, \cite{bayesian_learning_mpc}.

Even if some of these works have pushed forward research in the field of collaborative legged robots, they all share one major drawback from the
viewpoint of this paper: the collaboration is always carried out with an active manipulator, which brings additional complexity and failure points into the interaction task. In fact, active manipulators are costly since they require the presence of multiple actuators, and are prone to failure in case of fall given the intrinsic rigidity of the mechanisms.
%
Furthermore, the use of an actuated arm can significantly reduce the carrying capabilities of a legged robot, since the arm itself and its actuators usually account for most of the robot's payload capacity. For example, the arm mounted on the earlier version of the ANYmal robot weighed more than 40\% of its payload~\cite{bellicoso19icra}. Nowadays, actuators that perform torque control are becoming stronger and lighter, but still scale up significantly to carry high payloads\footnote{Here defined as above 30kg, which is nearly half of the global average weight of a person \cite{Walpole2012}.}. The actuated arm also requires a whole-body controller that takes into account the additional degrees of freedom of the arm for motion planning and control~\cite{minniti2022}, increasing the complexity of the control scheme.

\begin{figure}
    \centering
    \includegraphics[width=\columnwidth]{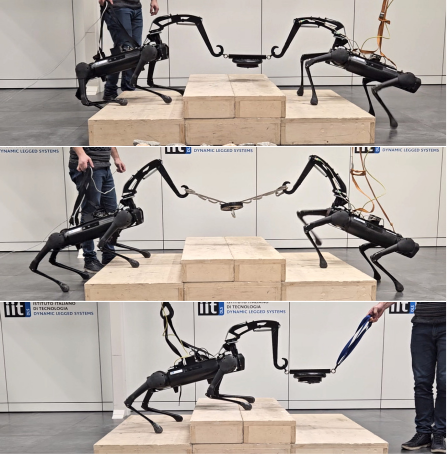}
    \caption{Experimental scenarios to assess three different collaborative carrying tasks using PACC:
    on the top-figure, robot-robot CC with rigid payload coupling; on the mid-figure, robot-robot CC with non-rigid payload; and on the bottom-figure, human-robot CC with rigid payload coupling.}
    \label{fig:exp_scenario}
\end{figure}

For simpler tasks, such as general load transportation in construction sites or the transportation of accident victims in rescue missions, in which the main objective is simply to carry a payload and there is no need for grasping maneuvers, an actuated arm is not necessary. In fact, for Collaborative Carrying (CC) tasks, most of the works design a simpler mechanism to attach the payload to the robot~\cite{fawcett23icra,yang22ral,rssVincenti2023}.

Several previous works have presented two or more legged robots performing collaborative payload carrying~\cite{distributed_col_dd,kim2023}. Most of them place the load between the robots using a rigid mechanism, which implies additional holonomic constraints that need to be handled specifically by the locomotion controller. Furthermore, such mechanism is not well suited for statically or quasi-statically stable locomotion, in which each robot needs to adjust its base position with respect to the support polygon independently. For such cases, a flexible mechanism is required to link the robots and the load. 

Other works explored collaborative carrying tasks with flexible connections between the robots. Yang et al.~\cite{yang22ral}, for example, combined a centralized and decentralized planning for collaborative manipulation of multiple legged robots, exploiting a rope for pulling objects. However, the robots do not compensate for the load and are shown to perform only a trot gait in flat terrain. Furthermore, their design, being a rope, permits only the transportation of an object \textit{together} with other robots. 

For human-robot collaborative transportation tasks, one of the main challenges is designing a control framework that allows the robot to follow the human without knowing his/her intention. This can be performed by using haptic measurements or employing a motion-capture system that provides information about the human motion. Sirintuna et al.~\cite{sirintuna2023} formulate an adaptive control framework to handle objects with unknown deformability for human-robot co-transportation tasks. They combine a whole-body controller with an admittance controller based on a motion capture system for the human behavior, which hinders the application of the approach in real-world industrial scenarios. 
Xinbo et al.~\cite{xinbo2021} proposed a human-robot co-carrying framework that uses visual and force sensing to estimate the human motion and an adaptive impedance-based controller for trajectory tracking. However, real-world applications can be limited by the need for a calibration board for visual feedback of the human-hand motion. Furthermore, these two works focus on collaborative tasks with wheeled robots, which are not appropriate for rough terrain scenarios. 


In this paper, we propose a novel mechanism for robot-robot/human-robot interaction, with the aim of resolving the aforementioned limitations. Our prototype is a 3 degrees-of-freedom arm based on passive elements, such as springs, which modulate automatically the impedance of the interaction, and only need joint encoders to operate, rendering the design lightweight and more robust to accidental damage.

Regarding the locomotion controller, Model Predictive Control (MPC) has become a popular choice for legged robots due to the possibility of ensuring constraints and considering the terrain, optimizing footholds, and handling external disturbances. For legged robot control during collaborative tasks, either between two robots or between robots and humans, two main approaches are commonly used. The first one is the application of a centralized controller that takes into account the state of all robots and that models the shared payload as rigid bodies~\cite{rssVincenti2023, yandong21case, kim2022thesis}. The second one is the use of a decentralized controller with a load compensation mechanism in each robot to handle the joint payload~\cite{kim2023,sombolestan23iros,humand_humanoid_col}. The latter is easier to implement and has the advantage of being more reactive, which makes it more suitable for rough terrain navigation and human interaction due to the naturally unpredictable behavior of humans.

From the control point of view, we design a distributed controller for robot collaboration that does not require any centralized information
to carry out the designed task, enhancing its scalability in the presence of multiple robots. Based only on proprioceptive information coming from the passive arms, our controller enables the robot to collaborate seamlessly with
humans and other systems, enabling it to understand collaboration intention and safely traverse rough terrain during the interaction. To achieve this last point, we employed an MPC with a modified dynamic stability criterion, which takes into consideration the force exchanged during the interaction to enhance the robot's locomotion stability.

To summarize, the main contributions of this paper are:
\begin{itemize}
    \item A concept and first design of a passive-arm mechanism, used for
    collaborative carrying with multi-legged systems, that enhances locomotion
    on challenging surfaces;
    \item An MPC control formulation that incorporates the influence of the
    coupling effects between payload and robot introduced by the passive-arm impedance;
    \item Experimental assessment and validation of the proposed mechanism and
    controller on three collaborative carrying scenarios over different types of rough terrain.
\end{itemize}

The paper is organized as follows: in Sec. \ref{sec:passive_arm}, we introduce
the mechanical design of a Passive-Arm for high-payload Collaborative
Carrying (PACC), the concepts behind it, and its features.
The MPC-based locomotion control strategy, which makes use of PACC to stabilize
and drive the robot, is explained in Sec. \ref{sec:locomotion}. Experimental
results are shown in Sec. \ref{sec:results} and Sec. \ref{sec:conclusions}
concludes the paper.

\section{Passive Arm Design and Features} \label{sec:passive_arm}

\begin{figure}
    \centering
    \includegraphics[width=1\linewidth]{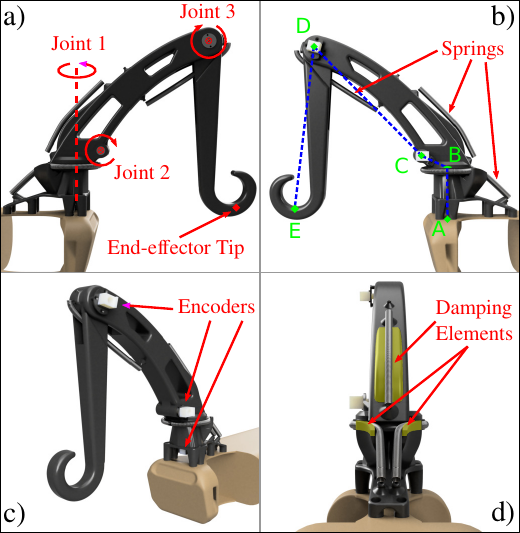}
    \caption{PACC mechanical design and components: a)~the arm has 3 revolute joints,
    arranged in a yaw-pitch-pitch configuration; b) the location of the springs and kinematics
    of the arm (see Table I for the data referring to the green dots). Note that joint 1 and 3 can also be equipped with 
    an antagonistic pair of springs instead of only one spring (as seen in the figure);
    c) each joint is equipped with an encoder to measure angular displacements; d) structural
    elements, highlighted in yellow, create friction forces against the spring extension and retraction to insert
    damping. The 4 subfigures show the arm in the joint zero-position configuration. Kinematic
    data related to the green dots are described in Table \ref{tab:kin_params}.}
    \label{fig:mech_design}
\end{figure}

The PACC proposed in this paper has 3 degrees-of-freedom (DoF)
and does not have joint actuators, see Fig.~\ref{fig:mech_design}.
It is composed of 3 revolute joints, in \textit{yaw-pitch-pitch} configuration,
where each joint torque is created by the extension and/or compression
of springs and other mechanical elements that introduce damping.

Different from previous designs used for collaborative carrying with quadruped
robots, we propose a structure that introduces intrinsic compliance between
the payload and the robot. Such compliance is fundamental to achieving safer
locomotion on very rough terrain where a trot does not represent the
ideal choice for the robot's gait or in case the robot must perform
a non-periodic gait and quasi-static or static locomotion. The core problem lies in the fact that, differently from the trot, in
quasi-static and static locomotion the robot needs to sway the trunk
to maintain its stability, and having a stiff connection between the robot
and the payload forces the sway (or the end-effector motion) of the robot and
its carrying companion (robot or person) to be synchronized. However, the
synchronization of sways or end-effector motions on rough terrain cannot be
guaranteed, considering that both carriers walk on a random irregular
surface, which tends to require different body attitudes and corresponding
footholds. In the remainder of this section, we describe the main elements involved
in the mechanical design and how the joint impedances are selected for the
intended object-carrying tasks.

\subsection{Mechanical Design} \label{sec:mech_design}

Apart from focusing on a sufficient arm workspace and a lightweight
design to maximize the robot carrying capability, the proposed design
has as core aspects its impact on the robot stability and on the
robot motion generation and control. For what regards the robot stability,
the kinematic design aims to reduce harmful disturbances on the trunk.
Therefore, in our passive arm, the tip (hook), where the payload is hooked, is
located at the level of the robot center-of-mass (CoM). During operation,
the arm hook height tends to work at a lower height than the repose
configuration, due to the gravitational forces from the payload. A lower
working height for the end-effector is beneficial because it impacts
less on the robot's Zero-Moment Point (ZMP) \cite{handbook_legged}
in case of strong horizontal forces, which are mainly present during
collaborative carrying using, e.g., cables or ropes (that get tensioned
to prevent the load from touching the ground). 

Regarding the aspects related to the robot's motion generation and control,
the design gets inspiration from the pendulum motion and dynamics, where
the third link is meant to behave mostly like a pendulum during the carrying
task. When the arm-tip is oriented toward the average payload motion, the most
relevant external interaction forces lie inside the \textit{pendulum plane}, i.e.,
the plane orthogonal to the third joint axis and that crosses the third link.
In this way, the angular position and the oscillation of the third link
w.r.t. to the gravity vector, can be exploited to estimate the average payload
motion, to give motion commands to the robot, and to estimate future external
forces (as described in detail in Sec. \ref{sec:guidance}
and Sec. \ref{sec:controller}).

The size of the third link is chosen so that its pendular displacement allows
the robot to move its ZMP inside the support polygon without experiencing
critical external forces from the interaction. The size of the second link,
in turn, is chosen so that the expected pendular motion of the third link
can occur without colliding with the robot's head (similarly if instead
installed in the back of the robot), and to prevent collisions with the
legs in case of large displacements of the first joint. All the kinematic
parameters are described in Tab. \ref{tab:kin_params}. In total, the
passive arm weighs about 1.3 kg, comprising: base-link (0.366kg), first-link
(0.162kg), second-link (0.497kg), and third-link (0.258kg).



\begin{table}[h]
\caption{PACC linear and angular kinematic parameters.}
\label{tab:kin_params}
\centering
\begin{tabular}{ccccccc}
\toprule
\multicolumn{4}{c}{\textbf{Distances [m]}}
& \multicolumn{3}{c}{\textbf{Angles [Degree]}} \\
\cmidrule(rl){1-4} \cmidrule(rl){5-7}
{$\overline{AB}$} & {$\overline{BC}$} & {$\overline{CD}$} & {$\overline{DE}$}
& {$\widehat{ABC}$} & {$\widehat{BCD}$} & {$\widehat{CDE}$}\\
\midrule
0.082& 0.056 & 0.271 & 0.277
& 117.6 & 164.7 & 51.6\\
\bottomrule
\end{tabular}
\end{table}

Although the design targets CC, its application can also be extended to
tasks performed alone, like dragging objects using ropes or carrying
objects that can be hooked. 

\subsection{Joint Impedance Selection} \label{sec:impedance_sel}

The design of the joint stiffness and damping elements has three main criteria:
1) the positioning of the end-effector in a convenient location inside
the arm workspace during the carrying task; 2) to serve as a means to
estimate the forces on the arm tip/hook; and 3) to shape the natural
dynamics of the subsystem formed by the payload and the arm in a way it can
be exploited by the locomotion controller and reduce the disturbance to the robot's
base motion. 

The impedance of the third joint can assume two configurations:
1) antagonistic arrangement, when antagonistic springs are installed to
create a stiffer neutral joint position, which is intended for tasks
where the robot must carry or drag some payload alone; 2) asymmetric, when
only one spring is installed to create a joint pre-torque that is useful
to control the payload height from the ground when the payload is carried using
ropes. 

There is an interesting feature to be highlighted. In the case of pure
inertial payloads, and if the forces generated by the impedance of the third joint are relatively small in comparison to the payload inertial
effects, it is possible to estimate the oscillatory dynamics of the
payload without knowing the payload mass. That is why a pure pendular
motion is independent of mass and only dependent on the length of
the pendulum. This feature becomes very relevant to estimate future
external forces caused by the payload oscillation, and feedback them into locomotion controllers that work based on receding horizons (e.g. MPCs).

When the arm is unloaded, or when there is only vertical gravitational force
acting on the payload, the antagonistic pair of springs of the first joint stabilizes
the first joint at zero position (when the arm is oriented towards the trunk
longitudinal direction). However, during the carrying task, the impedance of
the first joint is compliant enough to let the arm orient toward the payload
motion. All three estimated arm joint impedances are described in Table
\ref{tab:joint_impedances}. Since the third joint does not have any designed
damping mechanism, the only source of damping comes from its shaft and it is
assumed to be negligible. 

\begin{table}[h]
\caption{Measured joint impedances.}
\label{tab:joint_impedances}
\centering
\begin{tabular}{ccccccc}
\toprule
\multicolumn{3}{c}{\textbf{Stiffness [Nm/rad]}}
& \multicolumn{3}{c}{\textbf{Damping [Nms/rad]}} \\
\cmidrule(rl){1-3} \cmidrule(rl){4-6}
{Joint 1} & {Joint 2} & {Joint 3}
& {Joint 1} & {Joint 2} & {Joint 3}\\
\midrule
3.5& 8.47 & 2.75
& 0.26 & 1.43 & -\\
\bottomrule
\end{tabular}
\end{table}


\section{Locomotion Guidance and Control} \label{sec:locomotion}
In this section, we first describe how the velocity references
are generated using the passive arm for a robot that needs to act
as a follower. After that, we describe how the foot contacts are
planned and how the external forces at the end-effector are estimated, 
which are important elements of the proposed MPC formulation
presented at the end. 

\subsection{Locomotion Guidance} \label{sec:guidance}
\begin{figure}
    \centering
    \includegraphics[width=1\linewidth]{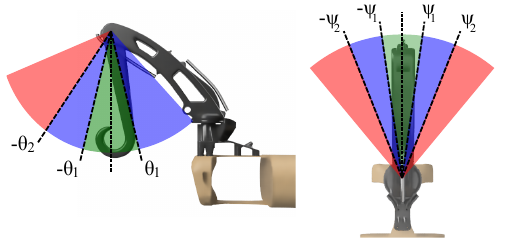}
    \caption{Velocity command zones according to the arm joint displacements: on the left,
    lateral view showing the angular ranges used to obtain the desired robot forward velocity,
    where $\theta$ is the orientation of Joint 3 with respect to the gravity vector;
    on the right, top view showing the angular ranges used to obtain the desired robot's
    heading velocity, where $\psi$ is the angular position of Joint 1.}
    \label{fig:arm_vel_commands}
\end{figure}

To drive the robot in a collaborative carrying task where a leader
(another robot or a person) is in charge of choosing the path,
we propose a simple approach using the angular displacements of
the passive arm to generate motion references. In synthesis,
the \textit{follower-robot}, or just \textit{follower}, is commanded in terms of forward
velocity (velocity along the longitudinal axis of its horizontal
frame \cite{barasuol13icra}) and heading velocity. The desired
forward velocity and heading velocity are computed according to
the displacement of Joint 3 and Joint 1,
respectively, as illustrated in Fig. \ref{fig:arm_vel_commands}.
There, the green range represents the
neutral angular range where the desired velocity is zero. The
blue range refers to the first velocity level and the red range
the second level (with increased velocity). Since such velocity
modulation is discontinuous, we apply a second-order low-pass
filter to smooth out the command signals. The values for the range limits,
defined by $\theta$ and $\psi$, and the velocities considered in
this paper are detailed in Sec.
\ref{sec:results}.

\subsection{Foot Trajectory Generation}
Each foot performs a cubic spline between foot contact positions based on the gait parameters and on the desired $x$ (longitudinal) and $y$ (lateral) body velocities. Thus, the nominal $xy$ coordinates of the footholds of the $i$th leg at the horizontal frame are defined as 
\begin{equation}
    \label{eq:footholdnom_xy}
    {\FootPosNomXY}_i = {\FootPosCenterH} + \frac{1}{2}\frac{\dutygait}{\stepfreq}\UserVfxyH + {\deltaFootPA}_i
\end{equation}
where ${\FootPosCenter}_i$ is the foot default position (\textit{home position}), $\dutygait$ is the ratio of the stance period and the step period, $\stepfreq$ is the step frequency, $\UserVfxyH$ is the desired $xy$ velocity defined in the horizontal frame; and ${\deltaFootPA}_i$ is a correction term to adjust the support polygon due to external forces at the end-effector
\begin{equation}
    \label{eq:foothold_deltaforce}
    \deltaFootPA = 
    {
    \Large
    \begin{bmatrix}
        \frac{\PAFeeHx\PAPeeHz - \PAFeeHz\PAPeeHx}{\massRobot\gravity - \PAFeeHz}&
        \frac{\PAFeeHy\PAPeeHz - \PAFeeHz\PAPeeHy}{\massRobot\gravity - \PAFeeHz}
    \end{bmatrix}}\Transp\text{,}
\end{equation}
where ${\PAFeeH}$, ${\PAPeeH}$ are the force and the position of the end-effector in the horizontal frame; $\massRobot$ is the total mass of the robot, and $\gravity$ is the gravity value.

The $\FootPosNomZ$ component of the foothold is computed based on the previous lift-off height $\FootPosLiftOffZ$, the desired linear velocity, and the desired angular rate $\CoMAngRateDesired$ of the robot's CoM
\begin{equation}
    \label{eq:footholdnom_z}
    {\FootPosNomZI} = {\FootPosLiftOffZ}_i -  \frac{\dutygait}{\stepfreq}
    \UserVz\text{,}
\end{equation}
where $\UserVz$ is the desired linear velocity for the base projected in the direction of the locally estimated terrain inclination. The next footholds in the world frame are predicted based on the desired velocity and remaining swing time of the $i$th leg ${\DeltaSwingtime}_i$
\begin{equation}
    \label{eq:footholdnom}
    {\FootPosNom}_i = \rotMatHTOW{\FootPosNomHI} + \CoMposlin + {\DeltaSwingtime}_i\rotMatHTOW\UserVf\text{,}
\end{equation}
where $\rotMatHTOW$ is the rotation from horizontal to world frame, ${\FootPosNomHI}$ is the nominal foothold with components defined by \eqref{eq:footholdnom_xy} and \eqref{eq:footholdnom_z}; and $\CoMposlin \in \Real^3$ is the position of the robot's CoM in the world frame.

\subsection{Locomotion Controller} \label{sec:controller}

\subsubsection{End-effector Force Estimation}
Let us consider the dynamic model of the passive arm subject to an external force in the end-effector
\begin{equation}
    \label{eq:passive_arm_model}
    \PAinertiaM\PAjointAcc + \PAV = \PAtauG + \PAtauSpring + \PAtauDamper + \PAJee\Transp\PAFee\text{,}
\end{equation}
where $\PAjoint \in \Real^3$ is the joint position of the passive arm, $\PAinertiaM \in \Real^{3\times3}$ is the inertia matrix, $\PAV \in \Real^3$ captures the Coriolis effect, $\PAtauG \in \Real^3$ is the gravitational torque, $\PAtauSpring \in \Real^3$ is the torque from the joint stiffness,
$\PAtauDamper \in \Real^3$ is the torque from the joint damping, $\PAJee \in \Real^{3\times3}$ is the end-effector Jacobian, and $\PAFee \in \Real^3$ is the force applied at the end-effector. Due to the low inertia of the third link, and assuming small joint velocities and small accelerations for the first and second joints, the terms $\PAinertiaM\PAjointAcc$ and $\PAV$ can be neglected. Hence, the force is computed as
\begin{equation}
    \label{eq:est_ext_force_ee}
    \KFFeeEst = -(\PAJee\Transp)^{-1}(\PAtauG + \PAtauSpring + \PAtauDamper) \text{,}
\end{equation}
where $\PAtauSpring$ and $\PAtauDamper$ are computed as
\begin{equation}
    \PAtauSpring = -\PASpringCoef (\PAjoint - {\PAjoint}_0)\text{ and }\PAtauDamper = -\PADamperCoef \dot{\PAjoint} \text{,}
\end{equation}
where $\PASpringCoef$ and $\PADamperCoef$ are, respectively, the diagonal
matrices of spring and damping coefficients from Tab. \ref{tab:joint_impedances},
and ${\PAjoint}_0$ is the vector of joint positions at the springs' equilibrium length. 

Thus, the equivalent force and torque in the robot's CoM described in the world frame are
\begin{gather}
    \label{eq:force_torque_ext}
    \begin{aligned}
        \ForceExt &= \rotMatBTOW\KFFeeEst\text{,}&
        \TorqueExt &= \rotMatBTOW \PAPee^B \times \KFFeeEst\text{,}
    \end{aligned}
\end{gather}
where $\PAPee^B$ is the end-effector position in the base frame.

\subsubsection{MPC}
A nonlinear MPC is formulated to compute ground reaction forces (GRFs) and foot positions to track a desired CoM's trajectory.  Similar to \cite{nmpc_angelo}, a
Single Rigid Body Dynamics (SRBD) model is considered
\begin{gather}
    \label{eq:srbd_model}
    \begin{aligned}
    &\massRobot(\ddot{\CoMposlin} - \gravityArray) = \sum_{i=1}^4 \contactFlag_i\Grf_i + 
    \dss_{lin}\text{, }\ \ \
    %
    \dot{\CoMposang} =  \TransOmegaToEuler(\CoMposang)\CoMvelang\text{,}
    \\
    &\inertiaRobot\dotCoMvelang + \CoMvelang\times \inertiaRobot{\CoMvelang}  = \sum_{i=1}^4 \contactFlag_i\contactPoint_i \times \rotMatBTOW\Grf_i + \dss_{ang}\text{,}\\
    %
    \end{aligned}
\end{gather}
omitting the swing legs' dynamics. In \eqref{eq:srbd_model},  $\CoMposlin \in \Real^3$ is the position of the CoM in the world frame, $\gravityArray \in \Real^3$ is the gravity vector; $\inertiaRobot^{3\times3}$ is the inertia tensor of the robot given in the base frame; $\contactPoint_i \in \Real^3$ is the $i$th foot position given in the base frame; $\Grf_i \in \Real^3$ is the GRF of the $i$th foot, and $\contactFlag_i$ is the contact state variable of the $i$th foot; $\TransOmegaToEuler(\CoMposang) \in \Real^{3\times3}$ is the transformation matrix between the angular velocity $\CoMvelang\in \Real^3$ in the base frame and the rate of the Euler angles $\dot{\CoMposang}$; $\dss_{lin}$ and $\dss_{ang}$ are, respectively, the linear and angular components of the external disturbance force $\dss$ $\in \Real^6$ applied in the CoM. In order to optimize the footholds location $\contactPoint_i$, we augment the input space of the controller by an additional variable $\FootVelOpt \in \Real^{12}$, obtaining a simplified model for the $i$th foot position dynamics as 
\begin{equation}
    \label{eq:foot_model}
    \FootVel_i = (1 - \contactFlag_i) \FootVelOpt_i\text{.}
\end{equation}

Considering as external disturbances only the forces acting at the end-effector and that
they can be estimated using Eq. \ref{eq:est_ext_force_ee} and Eq. \ref{eq:force_torque_ext},
it is possible to include in the prediction model an approximation for the evolution of such forces
due to the movement of the base. To do so, we simplify the modeling assuming that both the
end-effector and the base can move independently from each other and that the end-effector
positions along the horizon are known. This way, the distance between them along the horizon is given 
by:
\begin{equation}
\Delta_{er} = \PAPeeHat - \CoMposlin
\end{equation}
where $\PAPeeHat$ is the expected end-effector position along the horizon. If all forces between
the end-effector and the base are assumed only caused by the deformation of the springs, the
disturbance and its evolution are given as:
\begin{equation}
    \dss = \begin{bmatrix}
        \mathrm{f}^0_{\mathrm{ext}_x} + \BarPASpringCoefScalar{}_x (\Delta_{{er}_x} - \Delta^0_{{er}_x}) \\
        \mathrm{f}^0_{\mathrm{ext}_y} + \BarPASpringCoefScalar{}_y (\Delta_{{er}_y} - \Delta^0_{{er}_y})\\
        \mathrm{f}^0_{\mathrm{ext}_z}\\
        \tau^0_{{\mathrm{ext}}_x}\\
        \tau^0_{{\mathrm{ext}}_y}\\
        \tau^0_{{\mathrm{ext}}_z} - ||\Delta_{er}||^2\BarPASpringCoefScalar{}_y (\psi-\psi^0)
    \end{bmatrix}
\end{equation}
\begin{equation}
    \label{eq:ext_disturbance_model}
    \dot{\dss} = \begin{bmatrix}
        -\BarPASpringCoefScalar{}_x \CoMvellinx&
        -\BarPASpringCoefScalar{}_y \CoMvelliny&
        0&
        0&
        0&
        -||\Delta_{er}||^2\BarPASpringCoefScalar{}_y\YawVel
    \end{bmatrix}
\end{equation}
where the upper-index $0$ refers to initial measurements (considered
constant along the horizon) and 
\begin{gather}
\begin{aligned}
    \label{eq:fext_passive_arm}
    &\BarPASpringCoef = (\PAJee\Transp\rotMatWTOB)^{-1}\PASpringCoef(\rotMatBTOW\PAJee)^{-1} = 
    \mathrm{diag}(\BarPASpringCoefScalar{}_x,\BarPASpringCoefScalar{}_y,\BarPASpringCoefScalar{}_z)
\end{aligned}
\end{gather}

For ensuring dynamic stability, the ZMP is constrained to be inside the support polygon, e.g. the convex-hull generated by the leg of the robot in stance, which is defined based on the momentum associated with the passive arm end-effector~\cite{handbook_legged} as
\begin{gather}
    \label{eq:zmp_def}
    \begin{aligned}
        \xZMP = \frac{\massRobot\gravity \CoMposlinx - \CoMposlinz \massRobot\ddot{\CoMposlinScalar}_x - \PAPeex d_z + \PAPeez d_x}{\massRobot\gravity - d_z}\text{,}\\
        \yZMP = \frac{\massRobot\gravity \CoMposliny - \CoMposlinz \massRobot\ddot{\CoMposlinScalar}_y - \PAPeey d_z + \PAPeez d_y}{\massRobot\gravity - d_z}\text{,}
    \end{aligned}
\end{gather}
where $\CoMposlinScalar_j$, ${\PAPeeScalar}_j$, $d_j$ are the components of $\CoMposlin$, $\PAPee$, and $\dss$, respectively, with $j \in \{ x,y,z\}$. 

Finally, the state, control, and disturbance vector are defined as
\begin{gather}
    \label{eq:x_u_d_def}
    \begin{aligned}
    \xss &=
    \begin{bmatrix}
        \CoMposang\Transp &
        \CoMposlin\Transp &
        {\CoMvelang}\Transp &
        \dot{\CoMposlin}\Transp &
        \FootPos\Transp &
        \dss\Transp
    \end{bmatrix}\Transp\text{,}\\ 
    \uss &=
    \begin{bmatrix}
        \Grf\Transp & \FootVelOpt\Transp
    \end{bmatrix}\Transp\text{, }
\end{aligned}
\end{gather}
with $\xss \in \Real^{30}$ and $\uss \in \Real^{24}$. 

Given a periodic contact sequence that defines $\contactFlag$ along the prediction horizon $\Ny$, an optimization problem is formulated to track a desired state trajectory $\xref$
\begin{align}
    \label{eq:opt_mpc}
    &\underset{\Xopt_k,\Uopt_k}{\text{min}} &&
    \sum_{i=0}^{\Ny}\|{\xref}{}_{k+i} - \xss{}_{k+i}\|^2_{\Qfc}  + \sum_{i=0}^{\Ny-1}\|\uss{}_{k+i}\|^2_{\Rfc} + \|\boldsymbol{\epsilon}_{k+i}\|^2_\rho \\
    &\textit{s.t.}&& \xss_k = \xss_0\text{,}
    \\&&&
    \label{eq:model_function}
    \xss_{k+i+1} = \boldsymbol{g}(\xss{}_{k+i},\uss{}_{k+i},\contactFlagVector{}_{k+i})\text{,}
    \\&&&
    \label{eq:fc_contraint}
    \hFriction(\uss{}_{k+i}) \leq \ZeroMat\text{,}
    \\&&&
    \label{eq:zmp_contraint}
    \hZmp(\xss{}_{k+i}, \uss{}_{k+i}) + \boldsymbol{\epsilon}_{k+i} \leq \ZeroMat\text{,}
\end{align}
where $\Xopt_k$ and $\Uopt_k$ are the state and control actions along $\Ny$; $\xss_0$ is the current state using the values of $\ForceExt$ and $\TorqueExt$ given by \eqref{eq:force_torque_ext}; \eqref{eq:model_function} is the prediction model defined by \eqref{eq:srbd_model}, \eqref{eq:foot_model}, and \eqref{eq:ext_disturbance_model}; \eqref{eq:fc_contraint} comprehends the friction constraint defined by a pyramidal approximation and maximum force constraint~\cite{nmpc_angelo}; \eqref{eq:zmp_contraint} constrains the ZMP \eqref{eq:zmp_def} to be inside the support polygon defined by the leg in stance at time $k+i$, and $\boldsymbol{\epsilon}$ is an additional slack variables employed to avoid a potential infeasibility of the MPC. These variables, in fact, permit a small violations of the ZMP constraints instead of incurring in a failure of the MPC solver, and are heavily penalized in \eqref{eq:opt_mpc} by using the weight $\rho$. Finally, $\Qfc$, $\Rfc$, are diagonal weighting matrices that define the importance of the tracking error for each robot's state and input. 

The MPC\footnote{https://github.com/iit-DLSLab/Quadruped-PyMPC/} has been implemented using Acados~\cite{acados}, an open-source Sequential Quadratic Programming solver tailored for optimal control, employing HPIPM~\cite{HPIPM} as a quadratic solver.

\section{Results} \label{sec:results}
In this section, we present the results from the experimental tests
we made to demonstrate PACC with two quadruped robots (Unitree's Aliengo, 21kg mass).
We designed three different task scenarios involving robot-robot (RR) and
human-robot (HR) collaboration considering a \textit{leader-follower} motion coordination.
In all the scenarios, the robots are required to walk on a two-stage circuit
composed of stair-like obstacles and rocks. To do so, each robot executes
a periodic crawling gait with a step frequency of 0.5 Hz and a duty factor
equal to 0.8 (ratio computed as the period a foot stays in contact with
the ground over the step period). The locomotion is controlled by the
MPC formulation described in Sec. \ref{sec:controller} where the desired
margin for each robot's ZMP is set to 4 cm, the prediction horizon is $\Ny = 15$,
and the discretization time of the model \eqref{eq:model_function} is
set to $0.04$ seconds.

The first two demonstration scenarios involve robot-robot CC and the third
one human-robot CC, but with different types of carried payload, as shown in
Fig. \ref{fig:exp_scenario}. The leader-robot is commanded by means of an
operator computer. The follower-robot, in turn, implements the strategy
explained in Sec. \ref{sec:guidance} with the following parameters:
$\theta_1 = 10^o$, $\theta_2 = 25^o$, $\psi_1 = 10^o$, $\psi_2 = 20^o$,
$V_1= 0.1~m/s$, $V_2= 0.2~m/s$, $\dot\psi_1 = 0.3~rad/s$ and
$\dot\psi_2 = 0.4~rad/s$. In the remainder of this section, we provide more details about
each scenario.

\subsection{RR collaborative carrying of a rigid payload}

\begin{figure}[h!]
\centering
    \includegraphics[width=\columnwidth]{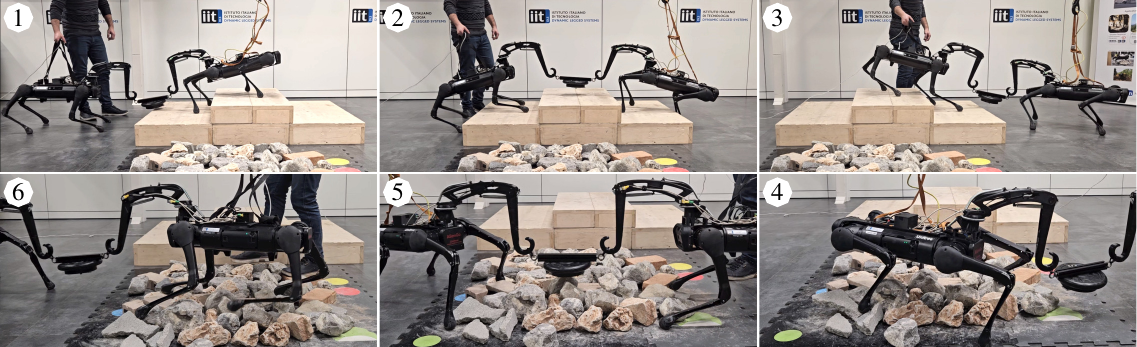}
    \caption{Sequence of numbered snapshots from experimental tests of collaborative carrying
    on a stair-like obstacle and rough terrain. The two quadruped robots are endowed with the
    passive arm to carry a 7kg payload connected to a stiff bar in a leader-follower manner.
    From snapshots 1 to 6, the robots walk up and down the stairs and turn around to cross
    the rocks. The stair obstacle is composed of pallets that are 55cm in depth and 16cm (bottom pallets)
    and 13cm (top pallet) in height.}
    \label{fig:rr_cc_rigid_payload}
\end{figure}

\begin{figure}[h!]
\centering
    \includegraphics[width=0.925\columnwidth]{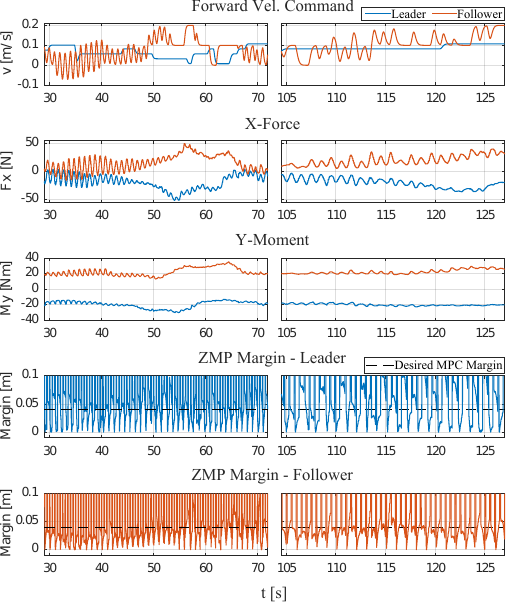}
    \caption{Plots for the experimental tests with two robots carrying a 7kg payload attached
    to a rigid bar. The left column refers to the period in which the robots walk on the stairs while
    the right column when traversing the rocks. From top to bottom, the first plot shows the
    forward velocity commands for each robot. The following two plots show the estimated
    forces along the longitudinal axis and moments on the sagittal plane of the trunk horizontal
    frame. The last two plots show the ZMP margin for each robot and the dotted line is the corresponding margin constraint.}
    \label{fig:data_cc_rr_rigid_payload}
\end{figure}

In this CC task, the robots are commanded to traverse the stairs and rocks
carrying a 7kg payload connected to a rigid bar attached on both
sides to the tip of each passive arm. Since the slack of the loops connecting
each arm to the bar is negligible, we consider this arrangement as a rigid
coupling that imposes a distance constraint between each arm tip. A sequence
of snapshots from the experimental test, showing both robots completing the
two-stage circuit, is depicted in Fig. \ref{fig:rr_cc_rigid_payload}. In
Fig. \ref{fig:data_cc_rr_rigid_payload} we show relevant data from the
experiment where we highlight aspects related to the robot commands,
interaction forces, and locomotion stability. To improve readability,
Fig. \ref{fig:data_cc_rr_rigid_payload} has two columns, where the left
column refers to the data when the robots are crossing the stairs, and
the right column when traversing the rocks.

The first plot of Fig. \ref{fig:data_cc_rr_rigid_payload} shows the guidance
strategy modulating the velocity command of the \textit{follower} to keep
the distance to the \textit{leader}. In the second plot, we see the interaction
forces along the longitudinal axis of each robot and, as expected, the average
force is negative for the \textit{leader} and positive for the \textit{follower}.
Note the relevant oscillations caused by the pendulum effect and the rigid
payload coupling arrangement. Such oscillations have a period of approximately 1s,
which is very close to the value of $2\pi\sqrt{l/g}$ ($l = \overline{DE}$),
validating the concepts behind the arm design. Despite the strong oscillations and variation
of the external forces, the MPC was able to make use of the estimated forces
for compensation and produce the needed trunk sway motion to keep the
ZMP inside the support polygon for both robots. This fact can be verified
in the two plots on the bottom. It is important to note that the average
ZMP margin is slightly lower for the \textit{follower} since it needs to
be reactive to follow the \textit{leader}.


\subsection{RR collaborative carrying of a deformable payload}
\begin{figure}[h!]
\centering
    \includegraphics[width=\columnwidth]{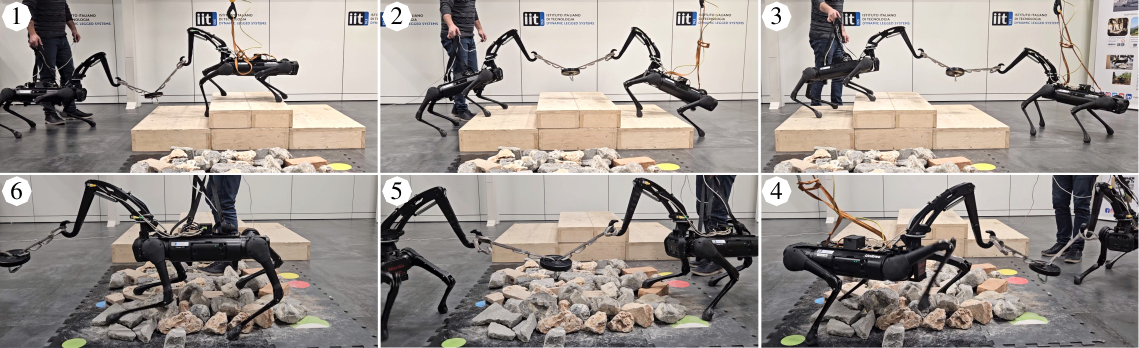}
    \caption{Sequence of numbered snapshots from the RR collaborative carrying experiment
    transporting a deformable 2kg payload.}   
    \label{fig:rr_cc_deformable_payload}
\end{figure}

\begin{figure}[h!]
\centering
    \includegraphics[width=0.925\columnwidth]{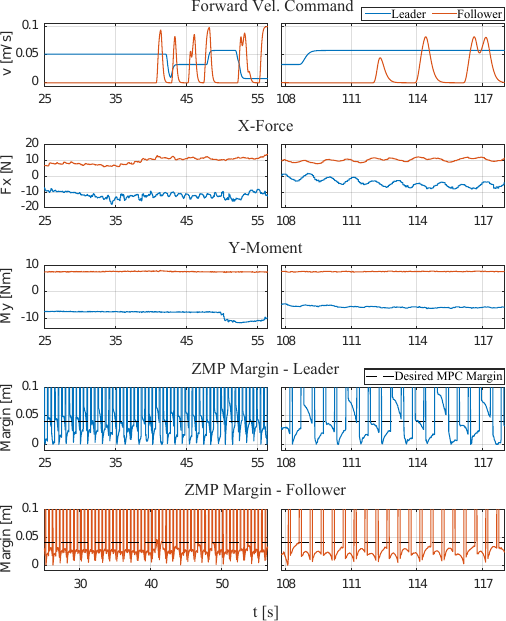}
    \caption{Plots for the experimental tests with two robots carrying a payload attached to a rope.
    The order and content description of the plots follow as the ones in Fig.~\ref{fig:data_cc_rr_rigid_payload}.
    }
    \label{fig:data_rr_cc_deformable_payload}
\end{figure}

Differently from the previous task, in this one, the two robots are required
to walk in the same scenario but carry a 2kg payload that is not
attached to the arms by means of a stiff element. Instead, the payload
is connected to the arms using a 60cm rope (30cm from the payload to
each arm tip). Therefore, we here use the terminology \textit{deformable payload}.
The main difference in this task is that the robots need to keep a minimum
of tension in the rope to prevent the payload from touching the ground. To do
so, we exploit the intrinsic arm impedance and shift $\theta_1$ and $\theta_2$
by 10 degrees. As for the first task, the results from this experimental
test are shown in Fig. \ref{fig:rr_cc_deformable_payload} and Fig.
\ref{fig:data_rr_cc_deformable_payload}. From the figures we can see that
the robots manage to complete the circuit, maintaining the locomotion stability
while keeping the tension of the rope sufficient to have the payload distant
from the ground. Note that the dynamics of the interaction forces are different
w.r.t. to the previous task, where the oscillations are less present and
the robot velocities are more constant.

\subsection{HR collaborative carrying of a rigid payload}
\begin{figure}
\centering
    \includegraphics[width=\columnwidth]{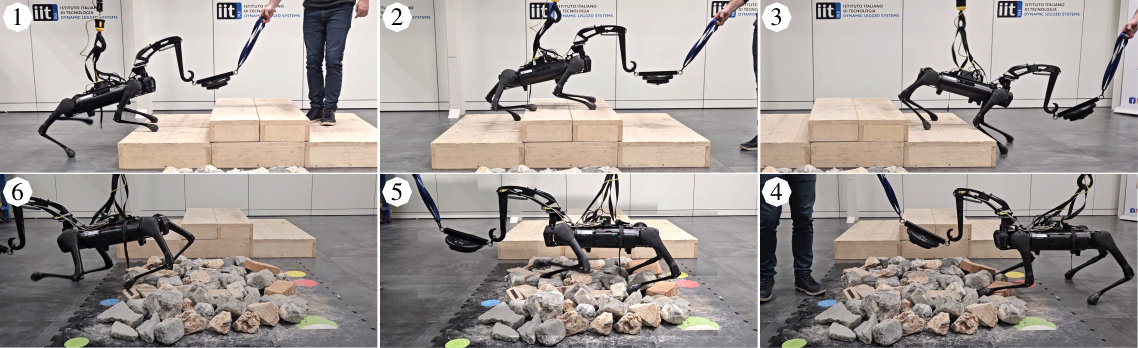}
    \caption{Sequence of numbered snapshots from the HR collaborative carrying experiment
    transporting a rigid 7kg payload.}
    \label{fig:hr_cc_rigid_payload}
\end{figure}

\begin{figure}
\centering
    \includegraphics[width=0.95\columnwidth]{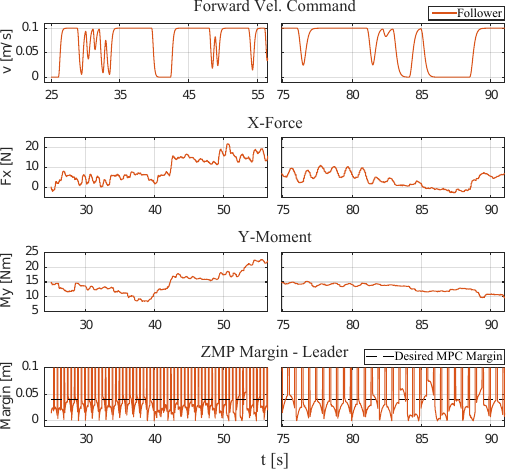}
    \caption{Plots for the experimental tests where a person (the leader)
    and a robot (the follower) carry together a rigid 7kg payload. The order
    and content description of the plots follow similarly as the ones in
    Fig.~\ref{fig:data_cc_rr_rigid_payload}.}
    \label{fig:data_hr_cc_rigid_payload}
\end{figure}

In this last task, we assess the robot's performance and analyze the
interaction forces during an HR collaborative carrying, in which
the human acts as the leader, to transport a rigid 7kg payload.
Similar to the previous tasks, the results from this experimental
test are shown in Fig. \ref{fig:hr_cc_rigid_payload} and Fig.
\ref{fig:data_hr_cc_rigid_payload}.

Differently from the previous results, in this task the relevant
interaction forces present a hybrid behavior between the features
from the first test and some from the second. In other words, it
presents smaller oscillations than observed in the first experiment
and more intense interaction forces than observed in the second one.
This is because the payload is as heavy as in the first experiment but
the human arm tends to stabilize the payload motion.


\section{Conclusions} \label{sec:conclusions}

In this paper, we introduced the concept of endowing multi-legged
systems with a passive arm with intrinsic impedance, called PACC,
to tackle the problem of collaborative carrying of high payloads over rough terrains.
Its benefits comprise the increase of the robot's net payload,
by not installing heavy actuated arms, reduction of system cost,
and mechanical robustness against undesired robot collisions or falls.
We presented a conceptual and mechanical passive-arm design and an
MPC formulation that takes into account the arm features to control
the robot locomotion. We showed that the kinematic morphology of the
arm can be used to drive the robot in case of a CC in leader-follower
manner. Most importantly, we showed that the passive arm allows quadruped
robots to use gait sequences that are ideal for very rough terrain, but
that require robot's body sway to guarantee locomotion stability.
We validated the approach by performing robot-robot and human-robot
CC over stair-like obstacles and rough terrain, and carrying different
payloads under blind locomotion.

Our future work will concentrate on the optimization of the mechanical
design, on the improvement of the motion references based on the displacement
of the third arm joint, and in the inclusion of vision-based foothold
selection and motion planning to improve the robot's terrain traversability.


\section*{ACKNOWLEDGMENT}

We would like to thank Franco Seguezzo for his support with electronics, and
Giovanni Dessy and Jo\~ao C. V. Soares for their support with the experimental tests.


\bibliographystyle{./bibtex/IEEEtran} 
\bibliography{00_root}
\addtolength{\textheight}{-12cm}   
\end{document}